\documentclass[11pt,a4paper]{article}
\usepackage[margin=2.5cm]{geometry}
\usepackage[T1]{fontenc}
\usepackage[utf8]{inputenc}
\usepackage{lmodern}
\usepackage{amsmath,amssymb,booktabs,graphicx,array}
\usepackage[numbers,sort&compress]{natbib}
\usepackage{tikz}
\usetikzlibrary{arrows.meta}
\usepackage{url}
\usepackage[hidelinks]{hyperref}
\hypersetup{pdfauthor={Yu Yan},
 pdftitle={JevForest: Path Voting for Budgeted Feature Acquisition},
 pdfkeywords={active feature acquisition, bagging, information gain, Newton boosting, tree ensembles}}
\title{JevForest: Path Voting for Budgeted Feature Acquisition}
\author{Yu Yan\\University of Chinese Academy of Sciences}
\date{}
\begin{document}
\maketitle
\begin{abstract}
Choosing which information to observe is central to prediction under limited observation budgets. We study JevForest, a feature acquisition policy that aggregates path-dependent proposals from bootstrapped trees, weights them by global training information gain, and predicts from the acquired values with a shared masked classifier. An online implementation queries Jev for semantic answers selected by this policy. On small balanced held-out samples, four-question forest acquisition achieves accuracy $0.729$ on AG News ($n=48$), compared with $0.667$ for a static gain ranking and $0.583$ for random ordering. On TREC ($n=24$), the ordering reverses: forest accuracy is $0.667$, compared with $0.750$ and $0.833$. Asking all eight questions in one batch yields higher accuracy at lower measured cost and latency than four sequential forest queries; direct Jev classification matches the batch accuracy while costing less. Offline MiniBooNE experiments yield accuracy $0.845\pm0.010$ at ten features and $0.885\pm0.008$ at forty features over three jointly varying data and forest seeds (mean $\pm$ sample standard deviation). A companion Newton boosting implementation provides preliminary full-feature synthetic results. These exploratory findings establish a working Jev acquisition workflow but do not support a general advantage for path voting: its value depends on the task, predictor, and the distinction between question budgets and actual query costs.
\end{abstract}

\begin{center}
\small Code is available at \url{https://github.com/qzqdz/jevforest}
\end{center}

\section{Introduction}
Prediction with a limited observation budget requires a model to decide what to measure before deciding what to predict. A feature that is useful on average may be redundant after other observations, while a less informative feature may become useful in a particular part of the input space. Sequential feature acquisition addresses this dependence by conditioning each choice on the values already observed. The resulting prediction quality depends on both the acquisition policy and the classifier applied to the partial input.

Tree paths offer a direct way to condition a choice on previous observations. Each internal node associates a feature with branches for its possible values. When some features are already known, a path can be followed until another observation is required. An ensemble then supplies several candidate next observations. The central question in this study is how aggregation of these proposals affects prediction under a hard feature budget.

We examine JevForest, which combines proposals from bootstrapped information-gain trees. Each tree uses a fixed random feature subset, and the ensemble weights proposal counts by feature information gain measured over the training set. A shared masked classifier produces the final prediction. This construction separates three components that can otherwise be conflated: conditional traversal of tree paths, global weighting of candidate features, and classification from the acquired values. The separation also makes vote-rule ablations possible without replacing the downstream predictor.

The contribution is the formulation and exploratory evaluation of this acquisition policy in two settings. Offline MiniBooNE experiments examine how weighting conditional path proposals affects accuracy across observation budgets. An online text-classification study connects the policy to the Jev Decisions API: each selected feature is a semantic question whose answer is obtained from a real model call. Comparing sequential acquisition with static and random question orders, all-question batching, and direct classification tests whether fewer questions translate into useful accuracy, cost, and latency tradeoffs. Neither setting establishes superiority over existing acquisition methods.

JevForest builds on shared tree and representation utilities~\cite{jevtree2026repo,jevforest2026repo}. JevGBoost~\cite{jevgboost2026repo} provides a companion reference implementation of full-feature Newton prediction. Its role is to distinguish aggregation of acquisition proposals from aggregation of predictive corrections over a compatible frozen representation. Its synthetic evaluation is preliminary, and no integrated forest--boost model is evaluated. The main experimental conclusions concern JevForest's observation choices, not a joint acquisition-and-prediction system.

\section{Related work}
Random forests aggregate randomized trees to improve prediction~\cite{breiman2001forest}. JevForest uses bootstrap sampling and feature subspaces, but aggregates next-feature proposals during acquisition. Its feature subset is sampled once per tree, whereas the conventional random-forest construction samples candidate features at individual nodes. Its final labels come from an external masked predictor, so tree voting and label prediction have separate roles.

Feature-Budgeted Random Forest learns forests for prediction under an average feature-acquisition cost constraint~\cite{nan2015fbrf}. This is a close connection between tree ensembles and observation cost. Our setting instead imposes a hard per-instance feature-count budget and uses the forest to select inputs for a shared external predictor. These differences in objective and prediction architecture motivate a separate evaluation; they do not imply an advantage over budget-aware forest prediction.

Other active acquisition methods learn policies or information estimates from partially observed inputs. Covert et al.~\cite{covert2023dynamic} study dynamic feature selection through mutual information, Li and Oliva~\cite{li2021surrogate} use generative surrogate models, and Valancius et al.~\cite{valancius2024oracle} study acquisition-conditioned oracle guidance for nongreedy acquisition. The present policy uses conditional tree paths with fixed global weights. It does not estimate conditional mutual information online or claim to optimize the objectives of these methods. A comparative study would require common data partitions, cost definitions, predictors, and tuning resources.

AFABench emphasizes modular comparison of acquisition policies, predictors, datasets, and evaluation settings~\cite{schutz2025afabench}. It provides relevant infrastructure for a broader controlled evaluation of path voting. Our local synthetic Cube generator is distinct from the official CUBE-NM benchmark; the shared name does not establish benchmark equivalence.

Typed decision primitives provide a broader motivation for selecting information. TypeSafe's Jev interface supports multiple typed questions evaluated in parallel on the same state~\cite{typesafe2026jev}. Recent preprints examine Jev accuracy and calibration across 37 datasets~\cite{deusser2026benchmark} and wireless decision-making, where faster individual calls need not improve end-to-end latency~\cite{rahimi2026wireless}. These settings make the distinction between questions, calls, and complete workflows consequential. A feature count is therefore not an API-call count or a latency estimate. Our MiniBooNE and Cube experiments are offline and do not invoke Jev. The separate online study uses Jev through OpenRouter~\cite{openrouter2026jev} and measures question budgets, reported monetary cost, and end-to-end latency independently.

For full-feature prediction, gradient boosting builds additive models by fitting successive corrections~\cite{friedman2001boosting}. Second-order tree scoring introduces gradient and Hessian statistics with regularized leaf values~\cite{chen2016xgboost}. JevGBoost follows this established structure over multiway discrete partitions. Its particular growth rules differ from standard positive-gain stopping and per-leaf regularization, as detailed below.

\begin{figure}[tbp]
\centering
\begin{tikzpicture}[x=1cm,y=1cm,
  font=\fontsize{9.5}{11.5}\selectfont,
  >={Stealth[length=1.8mm,width=1.3mm]},
  flow/.style={->,draw=jevink!65,line width=.5pt},
  box/.style={draw=jevblue!55,line width=.5pt,rounded corners=1.2pt,
    fill=jevblue!4,align=center,inner sep=0pt},
  boostbox/.style={box,draw=jevochre!65,fill=jevochre!5},
  note/.style={text=jevink!75,align=center,inner sep=0pt},
  tree/.pic={
    \draw[line width=.6pt] (0,.25)--(-.22,0) (0,.25)--(.22,0)
      (-.22,0)--(-.34,-.25) (-.22,0)--(-.10,-.25)
      (.22,0)--(.10,-.25) (.22,0)--(.34,-.25);
    \foreach \x/\y in {0/.25,-.22/0,.22/0,-.34/-.25,-.10/-.25,.10/-.25,.34/-.25}
      \fill (\x,\y) circle[radius=.036];
  }]
\definecolor{jevink}{HTML}{29343D}
\definecolor{jevblue}{HTML}{37667C}
\definecolor{jevochre}{HTML}{977136}
\path[use as bounding box] (0,0) rectangle (15.7,10.0);
\node[note,text width=15.3cm] at (7.85,9.75)
  {Shared representation: training-fitted discretization, with bins frozen at inference};
\draw[jevink!20,line width=.4pt] (.05,9.42)--(15.65,9.42);

\node[anchor=west,text=jevblue,text width=15.2cm,inner sep=0pt,
  font=\bfseries\fontsize{10}{12}\selectfont]
  at (.15,9.12) {(a) JevForest: feature acquisition};
\node[box,text width=2.1cm,minimum height=1.04cm] (masked) at (1.30,8.05)
  {Partial input\\$x_S$, mask};
\node[box,text width=3.4cm,minimum height=1.26cm] (forest) at (5.05,8.05) {};
\pic[jevblue] at (3.95,8.27) {tree};
\pic[jevblue] at (5.05,8.27) {tree};
\pic[jevblue] at (6.15,8.27) {tree};
\node[note,text=jevblue,text width=3.1cm] at (5.05,7.66) {Path proposals};
\node[box,text width=2.75cm,minimum height=1.04cm] (vote) at (9.1,8.05)
  {Weighted vote\\$V_j(x_S)\,G_j(D)$};
\node[box,text width=2.8cm,minimum height=1.04cm] (acquire) at (13.15,8.05)
  {Acquire feature $j^*$\\Update mask};
\draw[flow] (masked)--(forest);
\draw[flow] (forest)--(vote);
\draw[flow] (vote)--(acquire);

\coordinate (decision) at (13.15,6.91);
\draw[draw=jevink!65,line width=.5pt] (acquire.south)--(decision);
\fill[jevink!65] (decision) circle[radius=.04];
\draw[flow] (decision)--(1.30,6.91)--(masked.south);
\node[note,text width=4.8cm] at (6.30,7.17) {Repeat while budget remains};
\node[box,text width=3.25cm,minimum height=.95cm] (predictor) at (9.1,5.85)
  {Shared masked predictor};
\node[note,text=jevblue,text width=1.7cm] (label) at (5.05,5.85) {Label $\hat y$};
\draw[flow] (decision)--(13.15,5.85)--(predictor.east);
\node[note,anchor=west,text width=2.2cm,align=left] at (13.40,6.39)
  {Budget\\reached};
\draw[flow] (predictor.west)--(label.east);

\draw[jevink!20,line width=.4pt] (.05,5.14)--(15.65,5.14);

\node[anchor=west,text=jevochre,text width=15.2cm,inner sep=0pt,
  font=\bfseries\fontsize{10}{12}\selectfont]
  at (.15,4.82) {(b) JevGBoost: binary prediction};
\node[note,anchor=west,text=jevochre,text width=8cm,align=left]
  at (.15,4.38) {Training: labeled rows, sequential updates};
\node[boostbox,text width=2.8cm,minimum height=.94cm] (deriv) at (2.50,3.52)
  {Gradients $g_i$\\Hessians $h_i$};
\node[boostbox,text width=2.8cm,minimum height=1.22cm] (newtree) at (7.75,3.52) {};
\pic[jevochre] at (7.75,3.72) {tree};
\node[note,text=jevochre,text width=2.5cm] at (7.75,3.12) {Newton tree $f_t$};
\node[boostbox,text width=3.1cm,minimum height=.94cm] (margin) at (13.0,3.52)
  {Update margin\\$F_{t-1}\to F_t$};
\draw[flow] (deriv)--node[above=3pt,note,text width=1.3cm] {Fit} (newtree);
\draw[flow] (newtree)--node[above=3pt,note,text width=1.3cm] {Add $\eta f_t$} (margin);
\draw[flow] (margin.south)--(13.0,2.57)--(2.50,2.57)--(deriv.south);
\node[note,text width=4.4cm] at (7.75,2.31) {Next boosting round};

\node[note,anchor=west,text=jevochre,text width=8cm,align=left]
  at (.15,1.87) {Inference: full features, fixed trees};
\node[boostbox,text width=2.8cm,minimum height=.94cm] (full) at (2.50,.90)
  {Full input $x$};
\node[boostbox,text width=4.3cm,minimum height=1.50cm] (sumtrees) at (7.75,.90) {};
\pic[jevochre] at (6.45,1.20) {tree};
\pic[jevochre] at (7.75,1.20) {tree};
\pic[jevochre] at (9.05,1.20) {tree};
\node[note,text=jevochre,text width=4.1cm] at (7.75,.51)
  {$F_T(x)=F_0+\eta\sum_{t=1}^{T} f_t(x)$};
\node[boostbox,text width=3.1cm,minimum height=.94cm] (prob) at (13.0,.90)
  {Binary probability\\$p=\sigma(F_T)$};
\draw[flow] (full)--(sumtrees);
\draw[flow] (sumtrees)--(prob);
\end{tikzpicture}%

\caption{Two ensemble roles in JEV. (a) Bootstrapped trees propose unobserved features; gain-weighted voting selects acquisitions, and a shared predictor uses the budget-limited input. (b) During training, Newton trees update an additive margin. Inference evaluates the fixed ensemble on the full feature vector.}
\label{fig:ensemble_roles}
\end{figure}

\section{Methods}
\subsection{Observation budgets and prediction}
Let $D=\{(x_i,y_i)\}_{i=1}^{n}$ be a training sample with feature set $\mathcal F=\{1,\ldots,d\}$. An acquisition episode begins with $S_0=\varnothing$. A policy $\pi$ selects $j_k\in\mathcal F\setminus S_k$ from the observed values $x_{S_k}$ and their mask, then updates $S_{k+1}=S_k\cup\{j_k\}$. At budget $B$, a predictor $h_\theta$ returns $\hat y_B=h_\theta(x_{S_B},S_B)$. We consider equal feature costs and a hard constraint $|S_B|\leq B$ for each example. Evaluation measures the accuracy and macro-F1 of this policy--predictor pair. The forest voting rule is a heuristic for choosing observations, not the optimizer of a fitted acquisition-loss objective.

In the companion full-feature problem, a classifier maps the complete vector $x$ to a binary probability. Figure~\ref{fig:ensemble_roles} summarizes this distinction. Increasing the feature budget exposes more input to an acquisition predictor; increasing the number of boosting trees changes a classifier whose input is already available.

For continuous tabular features, both models use quantile boundaries fitted to training data. These boundaries remain fixed across trees and at inference, while categorical values retain their branch identities. In the following information-gain expressions, $D_{j=v}$ denotes the training subset whose discretized feature $j$ has value $v$. The acquisition predictor can still use the original observed values; discretization governs tree construction and traversal. Online Jev answers are thresholded into binary features before fitting or traversal.

\subsection{Training trees for feature proposals}
For class proportions $p_c$ in a sample, define entropy and feature gain by
\begin{align}
H_D(Y)&=-\sum_c p_c\log_2 p_c,\\
G_j(D)&=H_D(Y)-\sum_v\frac{|D_{j=v}|}{|D|}H_{D_{j=v}}(Y).
\end{align}
For each of $T$ trees, JevForest draws $n$ training rows with replacement and selects a feature subset. This subset remains fixed throughout tree growth. The square-root setting selects $\lfloor\sqrt d\rfloor$ features per tree. Information-gain splits are then fitted within the selected features, subject to depth and sample constraints. All trees use the same training-fitted quantile boundaries.

Given $x_S$, tree $t$ follows branches for observed features and proposes the first eligible unobserved feature encountered, denoted $q_t(x_S)$. It abstains if traversal reaches a leaf or encounters an unavailable branch. Different observed values can therefore produce different proposals, even though the trees remain fixed. Bootstrap sampling and feature subspaces provide variation among the candidate paths.

\subsection{Aggregating acquisition proposals}
The default rule counts proposals and weights each count by global training gain:
\begin{align}
V_j(x_S)&=\sum_{t=1}^{T}\mathbf{1}\{q_t(x_S)=j\},\\
j^*&=\underset{j\notin S,\;V_j(x_S)>0}{\arg\max}
       \;V_j(x_S)\bigl(G_j(D)+10^{-12}\bigr).
\end{align}
Ties favor the larger vote count and then the feature name in lexical order. If every tree abstains, the policy selects an eligible remaining feature from the global gain ranking. The observed set is updated after each acquisition until the budget is reached.

The factor $G_j(D)$ does not depend on the current mask or the node that proposed $j$. It is neither node-local gain nor conditional mutual information given $x_S$. Adaptation comes from the paths; the weights impose a fixed preference among their proposals. The two components can disagree: a frequently proposed feature with low global gain may lose to a less frequent proposal. Whether this tradeoff is useful can depend on the remaining information requirements and hence on the evaluation budget.

We compare this rule with plurality and out-of-bag (OOB) plurality. Plurality ranks raw proposal counts, breaking ties by global gain and feature name. OOB plurality replaces each unit ballot by
\begin{equation}
w_t=\frac{c_t+1}{n_t^{\mathrm{OOB}}+2},
\end{equation}
where $c_t$ is the number of correct predictions on the tree's $n_t^{\mathrm{OOB}}$ OOB rows. An empty OOB set gives weight $0.5$. These weights assess individual tree classification, which is not necessarily aligned with the value of their acquisition proposals.

\subsection{Classification from partial observations}
MiniBooNE uses a shared logistic predictor fitted on all training features. At inference, unobserved values are replaced by their training means. With standardized inputs, the positive-class output is
\begin{equation}
\hat p_\theta(x_S,S)=\sigma\left(b+\sum_{j\in S}\beta_j\frac{x_j-\mu_j}{s_j}\right),
\end{equation}
where $\mu_j$ and $s_j$ are training statistics and $\sigma$ is the logistic function. This is the classifier's output, not a claim that mean imputation yields the true conditional class probability.

Holding the predictor fixed isolates changes in the acquisition rule within the offline setting. The resulting scores describe the policy--predictor pair rather than the forest's native classification vote. Cube and the online text tasks use a shared matching-based majority predictor instead. It returns the majority label among training rows matching all acquired features. If no row matches, it progressively drops acquired features in ascending order of global information gain until a match is found; if none remains, it returns the training majority. This predictor remains fixed across the online acquisition baselines.

\subsection{Online acquisition of Jev answers}
For an input text $z$, each feature $j$ corresponds to a fixed semantic question $Q_j$. A request to Jev supplies $z$ as the state and asks $Q_j$ with explicit true and false criteria. The returned \texttt{noul} value $a_j\in[0,1]$ is converted to $x_j=\mathbf{1}\{a_j\geq0.5\}$. This threshold defines a binary representation; it is not a calibrated stopping rule. The forest chooses the next question from the current masked answer vector, the API supplies its answer, and the shared predictor produces a label after each update. Earlier answers condition local policy selection but are not appended to the text sent to Jev. Labels and training demonstrations are never included in the request state.

Training acquires all eight questions in one request per text. The resulting binary matrix and training labels fit the forest, global gain ranking, and matching predictor. At test time, sequential policies request one previously unasked question per step, while an all-question baseline requests all eight at once. The API model selector is \texttt{typesafe/jev-1.13}, served through OpenRouter's Decisions endpoint~\cite{openrouter2026jev}; every successful experimental response identifies snapshot \texttt{typesafe/jev-1.13-20260917}. Each online condition makes its own requests. Thus, the online comparison evaluates the complete acquisition workflow, including answer variability, rather than replaying a fixed matrix of cached test features.

The question budget counts requested semantic features. It does not equate requests with prices or assume that asking fewer questions is faster. Sequential acquisition can change the next question after observing an answer, but incurs repeated request overhead. Batching sacrifices this adaptation while sharing the input text across eight questions in one call. Monetary cost and elapsed time must therefore be measured alongside accuracy.

\subsection{Additive Newton prediction}
JevGBoost encodes two classes as $y_i\in\{0,1\}$ and initializes the margin $F_0$ with the logit of the training positive fraction, after clipping that fraction to $[10^{-6},1-10^{-6}]$. At round $t$, it computes logistic gradients and stabilized Hessians:
\begin{align}
p_i^{(t)}&=\sigma(F_{t-1}(x_i)),\\
g_i^{(t)}&=p_i^{(t)}-y_i, &
h_i^{(t)}&=\max\{p_i^{(t)}(1-p_i^{(t)}),10^{-6}\}.
\end{align}
For node indices $I$, let $G_I=\sum_{i\in I}g_i$ and $H_I=\sum_{i\in I}h_i$. The unconstrained regularized Newton value is $-G_I/(H_I+\lambda)$; the stored leaf value clips it to $[-5,5]$ by default and is zero when the denominator is at most $10^{-12}$.

A candidate feature partitions the node by its discrete values. At least two children are required, each satisfying the minimum sample count. For children $I_1,\ldots,I_K$, the split score is
\begin{equation}
\Delta_{\mathrm{impl}}=
\frac{1}{2}\sum_{k=1}^{K}\frac{G_{I_k}^{2}}{H_{I_k}+\lambda}
-\frac{1}{2}\frac{G_I^{2}}{H_I+\lambda}-\gamma.
\end{equation}
Three details qualify its relationship to regularized Newton optimization. The score uses unconstrained leaf optima although actual updates are clipped. The penalty charges $\gamma$ per split; a per-leaf penalty would charge $(K-1)\gamma$ when replacing one leaf with $K$ children. These conventions agree for binary splits or $\gamma=0$, but not for general multiway splits with nonzero $\gamma$.

Finally, if no positive-gain split is accepted, growth selects the first feature in the supplied order with a feasible partition. This fallback can accept zero or negative gain, allowing deeper interaction structure to be reached without guaranteeing an improvement in the regularized objective. Depth, sample constraints, and removal of previously used features along a path bound growth. An unseen branch uses a default leaf computed from the parent node statistics.

The fitted tree $f_t$ updates the margin and final probability as
\begin{align}
F_t(x)&=F_{t-1}(x)+\eta f_t(x),\\
p_T(x)&=\sigma\left(F_0+\eta\sum_{t=1}^{T}f_t(x)\right).
\end{align}
This model evaluates full feature vectors and has no acquisition policy. Its connection to JevForest is the shared representation and tree-based computation, not a common training objective.

\section{Experiments}
\subsection{Offline data and evaluation protocol}
The offline real-data experiment studies MiniBooNE particle identification~\cite{roe2005miniboone}, with 50 features, 2,000 training rows, and 500 test rows per run. Forests contain sixteen trees, with four quantile bins, depth limit eight, minimum-sample parameter eight, bootstrap sampling, and seven features per tree. Budgets are $5$, $10$, $20$, and $40$. Seeds $0$, $1$, and $2$ jointly vary data splitting and forest randomization. Accuracy and macro-F1 evaluate the shared masked logistic predictor after acquisition.

For each example, the multi-run evaluation acquires up to the largest budget and scores the resulting prefixes. Because the policy is deterministic given the current observations and does not condition on the target budget, these prefixes coincide with the sequences obtained from separate hard-budget episodes. This equivalence need not hold for randomized, stateful, or budget-aware policies.

The synthetic Cube task provides a small companion experiment. It contains 256 generated rows with five binary features and label $(f_0\mathbin{\oplus}f_1)\lor(f_0\land f_2)$, split into 179 training and 77 test rows. Forest acquisition uses sixteen trees, all features per tree, depth six, and minimum-sample parameter one. Boosting is evaluated separately with all features available. The task samples with replacement from finite binary support and therefore has limited value for assessing generalization to unseen feature combinations.

\subsection{Prediction quality across acquisition budgets}
Table~\ref{tab:forest} summarizes forest acquisition. Accuracy increases with budget in all three runs. At ten acquired features, mean accuracy is $0.845$ with sample standard deviation $0.010$; at forty features it is $0.885$ with standard deviation $0.008$. Macro-F1 rises from $0.615$ at five features to $0.850$ at forty, indicating that accuracy alone would give an incomplete account of low-budget class performance. Seed-zero accuracy and macro-F1 at forty features are $0.894$ and $0.866$.

\begin{table}[tb]
\centering
\caption{MiniBooNE acquisition performance. Three-run summaries show mean $\pm$ sample standard deviation over jointly varying split and forest seeds.}
\label{tab:forest}
\small
\begin{tabular}{rrrrr}
\toprule
Budget & Seed 0 accuracy & Seed 0 macro-F1 & Accuracy, 3 seeds & Macro-F1, 3 seeds\\
\midrule
5  & 0.798 & 0.659 & $0.785\pm0.021$ & $0.615\pm0.040$\\
10 & 0.856 & 0.797 & $0.845\pm0.010$ & $0.770\pm0.025$\\
20 & 0.878 & 0.840 & $0.869\pm0.013$ & $0.823\pm0.015$\\
40 & 0.894 & 0.866 & $0.885\pm0.008$ & $0.850\pm0.015$\\
\bottomrule
\end{tabular}
\end{table}

The across-run standard deviation uses denominator $3-1$ and combines split and model variation. It is not a standard error or an estimate of forest randomness alone. We also report seed-zero 95\% episode-bootstrap accuracy intervals from 1,000 resamples: $[0.762,0.834]$, $[0.826,0.886]$, $[0.848,0.906]$, and $[0.866,0.922]$ for increasing budgets. These intervals condition on the fitted model and resample test correctness without retraining. They exclude training uncertainty and do not quantify paired differences between methods.

\subsection{Budget-dependent effects of vote weighting}
The seed-zero ablation changes the vote rule while retaining the forest settings and reported data split. Table~\ref{tab:ablation} shows a local advantage for global gain weighting at budget ten: accuracy increases from $0.836$ under plurality to $0.856$. At budget twenty, the direction reverses, with accuracy decreasing from $0.882$ to $0.878$. The two rules coincide at budgets five and forty. OOB plurality gives lower accuracy at the three smaller budgets and the same accuracy at forty.

\begin{table}[tb]
\centering
\caption{MiniBooNE vote-rule ablation at seed zero, with seven features per tree and a shared masked logistic predictor.}
\label{tab:ablation}
\begin{tabular}{lrrrr}
\toprule
Vote rule & $B=5$ & $B=10$ & $B=20$ & $B=40$\\
\midrule
Plurality & 0.798 & 0.836 & 0.882 & 0.894\\
Global-IG weighted & 0.798 & 0.856 & 0.878 & 0.894\\
OOB plurality & 0.774 & 0.802 & 0.854 & 0.894\\
\bottomrule
\end{tabular}
\end{table}

The pattern does not support uniform dominance of gain weighting. It shows that a fixed preference among conditional proposals can help at one observation budget and hinder at another. Similarly, weighting trees by OOB classification performance does not yield better acquisition in this experiment. The results motivate separate analysis of proposal quality and classification quality, but do not establish the mechanism of the differences without acquisition traces.

The aggregation rule and square-root feature subset were selected after exploratory comparisons. These results are therefore subject to configuration-selection bias and require confirmation on an independent test partition. Per-example predictions were not retained for the ablation, preventing paired significance analysis of its small differences.

\subsection{Preliminary synthetic results}
Table~\ref{tab:cube} separates acquisition and full-feature prediction on Cube. Forest seed-zero accuracy is $0.688$, $0.909$, and $1.000$ at budgets one, two, and three. All five evaluated forest seeds reach $1.000$ at budget three, while budget-two accuracy has mean $0.839$ and sample standard deviation $0.139$. Saturation after three observations limits the task's ability to distinguish acquisition strategies.

\begin{table}[tb]
\centering
\caption{Cube acquisition and full-feature prediction. The rows represent separate evaluation settings and do not form a head-to-head comparison.}
\label{tab:cube}
\begin{tabular}{llr}
\toprule
Implementation & Setting & Accuracy\\
\midrule
Forest acquisition & Seed 0, budget 1 & 0.688\\
Forest acquisition & Seed 0, budget 2 & 0.909\\
Forest acquisition & Seed 0, budget 3 & 1.000\\
Newton boosting & Full features, 16 trees & 1.000\\
\bottomrule
\end{tabular}
\end{table}

The available boosting run achieves accuracy $1.000$ with sixteen trees. Its saved results omit the split seed, depth, and learning rate, so current defaults cannot establish the exact historical configuration. Moreover, the finite support permits extensive train--test overlap. In the current default seed-zero generation, 73 of 77 test rows have feature vectors seen during training. This count characterizes the current generator, not recovered metadata for that boosting run. The synthetic result demonstrates basic prediction behavior on a small task; it establishes neither generalization to unseen combinations nor an advantage over forest acquisition.

\subsection{Online text-classification protocol}
The online study uses four-class AG News~\cite{zhang2015character} and six-class TREC question classification~\cite{li2002question}. We draw 16 training examples per class, giving 64 AG News and 96 TREC examples. Held-out evaluation uses 12 examples per AG News class and four per TREC class, giving 48 and 24 test examples. The original train and test splits remain separate. Previously used pilot examples and duplicate normalized texts are excluded, including test texts matching selected training texts. The TREC test quota falls from five to four per class because of these exclusions. AG News is sampled from a convenience pool comprising the first 2,000 training and 500 test rows; TREC uses the complete cached source files. Sampling uses seed 20261007. These balanced subsets do not represent the full benchmark distributions.

Each task has eight fixed semantic questions. AG News questions concern government and diplomacy, armed conflict, sports contests, athletes and teams, financial markets, company activity, scientific research, and digital technology. TREC questions concern people or organizations, places, quantities, dates or durations, abbreviation expansion, definitions, procedures or reasons, and physical entities. The questions ask about these properties rather than the final class label. Their wording is fixed before held-out evaluation. Forests contain sixteen bootstrapped trees with two randomly selected features per tree, depth limit six, minimum-sample parameter four, and forest seed zero. Global-gain weighting and the matching predictor are fixed without test-based tuning.

We compare forest path voting with a static training-information-gain ranking and a seeded random question order. Each condition makes four sequential single-question requests per example, scoring the prefixes at budgets one, two, and four. Prefix scores reuse only earlier calls within that same trajectory. A fourth condition acquires all eight features in one batch and applies the same predictor. A fifth makes a single zero-shot \texttt{choice} request for the task label. The direct-choice condition uses Jev's own classification output and therefore provides a separate workflow reference, not a controlled acquisition-policy ablation.

Two workers process held-out examples concurrently; calls remain sequential within each trajectory. Training extraction initially used four workers and was reduced to two after transport failures. Example order and within-example method order are shuffled reproducibly. We measure client-observed end-to-end latency from acquisition through prediction, including network overhead, local processing, and any connection-handshake retries before an HTTP request is sent. Requests already sent are not automatically retried. We sum provider-reported USD charges; per-example evaluation costs exclude training feature extraction. All attempts remain recorded, and missing costs are treated as unknown. Accuracy uses the full intended test count, with unscored examples counted as failures. All held-out conditions completed successfully, so their cost and latency summaries cover the same 48 AG News or 24 TREC examples.

\subsection{Online accuracy and query costs}
Table~\ref{tab:live_accuracy} shows that path voting has different effects on the two tasks. At budget four on AG News, the forest classifies 35 of 48 examples correctly, compared with 32 for the static gain ranking and 28 for random ordering. On TREC, the corresponding counts are 16, 18, and 20 of 24. Thus, the online study does not reproduce a consistent advantage for the forest across tasks, even with a shared question set and downstream predictor.

\begin{table}[tb]
\centering
\caption{Accuracy from real sequential Jev calls. Budgets count semantic questions, with one request per question. All three methods use the same fitted masked predictor; prefixes share calls only within their own trajectory.}
\label{tab:live_accuracy}
\begin{tabular}{llrrr}
\toprule
Task & Acquisition order & $B=1$ & $B=2$ & $B=4$\\
\midrule
AG News & Forest & 0.417 & 0.604 & 0.729\\
 & Static gain & 0.458 & 0.479 & 0.667\\
 & Random & 0.417 & 0.438 & 0.583\\
\midrule
TREC & Forest & 0.333 & 0.500 & 0.667\\
 & Static gain & 0.333 & 0.500 & 0.750\\
 & Random & 0.417 & 0.542 & 0.833\\
\bottomrule
\end{tabular}
\end{table}

Exploratory paired bootstrap intervals resample example IDs within class strata 10,000 times, using seed 20261007. At budget four, the forest-minus-static accuracy difference is $0.0625$ on AG News, with 95\% interval $[-0.0625,0.1875]$, and $-0.0833$ on TREC, with interval $[-0.1667,0.0417]$. The forest-minus-random differences are $0.1458$ with interval $[0.0208,0.2708]$ and $-0.1667$ with interval $[-0.2917,-0.0417]$, respectively. These intervals condition on the fitted models and observed API answers. They neither include training or provider variability nor adjust for the multiple tasks, methods, and budgets examined; we make no confirmatory significance claim.

\begin{table}[tb]
\centering
\caption{Online workflow comparison. Cost is mean provider-reported USD per example, in units of $10^{-6}$ USD; time is median end-to-end seconds. Sequential methods use four questions in four requests. Batch uses eight semantic questions in one request. Direct uses one multiclass question and bypasses the masked predictor. Training extraction is excluded.}
\label{tab:live_cost}
\small
\begin{tabular}{llrrrrr}
\toprule
Task & Workflow & Questions & Calls & Accuracy & Cost & Time (s)\\
\midrule
AG News & Forest & 4 & 4 & 0.729 & 66.03 & 3.525\\
 & Static gain & 4 & 4 & 0.667 & 66.05 & 3.695\\
 & Random & 4 & 4 & 0.583 & 65.99 & 3.571\\
 & Batch & 8 & 1 & 0.771 & 38.07 & 0.863\\
 & Direct & 1 & 1 & 0.771 & 17.99 & 0.846\\
\midrule
TREC & Forest & 4 & 4 & 0.667 & 57.93 & 3.537\\
 & Static gain & 4 & 4 & 0.750 & 57.57 & 3.568\\
 & Random & 4 & 4 & 0.833 & 57.67 & 3.562\\
 & Batch & 8 & 1 & 0.917 & 36.66 & 0.811\\
 & Direct & 1 & 1 & 0.917 & 18.06 & 0.836\\
\bottomrule
\end{tabular}
\end{table}

Table~\ref{tab:live_cost} exposes the difference between limiting questions and reducing actual expense. Although the forest requests only four of eight semantic features, its repeated calls cost more and take longer than acquiring all eight in a batch on both tasks. Batch accuracy is $37/48$ on AG News and $22/24$ on TREC, exceeding the four-question forest. Direct Jev classification attains the same accuracy as batching at lower reported cost, with similar latency. The paired methods need not make the same errors, and the direct workflow changes the predictor as well as the question format. These measurements favor batching or direct classification for this pilot; fewer requested features alone do not demonstrate a practical efficiency benefit.

The run contains 1,168 successful experimental responses with distinct provider response identifiers: 160 for training extraction and 1,008 for held-out evaluation. Training incurred three additional transport failures with unknown billed cost. Successful training responses report a total of USD $0.005947284$, and held-out responses report USD $0.017668014$; a separate sanity probe reports USD $0.000013356$. These returned charges do not resolve the three unknown amounts. No held-out prediction is missing, and all 864 sequential acquisition steps were reconstructed from their recorded answers and fitted policies.

\section{Discussion}
\subsection{What the acquisition results establish}
The offline finding is that proposal weighting interacts with observation budget. Conditional paths offer a way to adapt feature choices without estimating a new conditional information score at each step. Global gain then changes the balance between agreement among trees and the marginal informativeness of a feature. The MiniBooNE ablation supports examining this balance at individual budgets rather than treating a single weighting rule as universally preferable. Online, the forest exceeds the sequential baselines at budget four on AG News but falls below them on TREC. Its four-question workflow is also less accurate, more costly, and slower than the all-question batch on both tasks. This distinction between acquisition accuracy and workflow efficiency is central to interpreting the method.

The use of a shared predictor is central to this interpretation. It makes the ablation a comparison of acquisition rules under one prediction model, while leaving the model's limitations in place. Mean-imputed logistic prediction need not exploit every interaction revealed by the policy. Thus, the measured curve describes a particular pairing of observation selection and classification, and a different masked predictor could change the ordering of policies.

\subsection{Scope and limitations}
The offline real-data evidence covers one task and three jointly varying seeds. Comparisons with earlier single-tree repository runs are excluded because preprocessing differed: removing 468 all-sentinel rows before splitting changed the forest evaluation sample. Common nominal seeds therefore did not produce a controlled baseline comparison. Claims of relative advantage on MiniBooNE require a new study with identical preprocessing and held-out examples for every policy. The online study adds two text tasks with matched held-out examples across methods, but uses small balanced samples and only one forest seed. It is an exploratory test of the live workflow, not a full benchmark evaluation.

The exploratory selection of vote rules further limits statistical interpretation. A validation partition should select hyperparameters before testing, and independently varied split and forest seeds should separate their sources of variability. Paired predictions and acquisition sequences would support uncertainty estimates for method differences and reveal when path proposals give way to global-ranking fallback. Unequal acquisition costs and broader datasets are also needed to assess settings beyond hard feature counts.

Boosting remains a preliminary companion study. Its fallback, clipping, and multiway penalty affect how closely tree growth follows a regularized Newton objective. Future ablations should isolate these choices against positive-gain stopping and per-leaf penalties on common binary-classification tasks. A larger evaluation also requires complete run configurations and established prediction baselines. The current synthetic result cannot support those conclusions.

A possible extension is to combine an acquisition forest with a predictor trained specifically for partial observations, including a boosted model. Such an integration would require explicit treatment of missing features and a new evaluation. The present study establishes neither this hybrid nor a shared acquisition-and-prediction objective; it identifies two distinct roles that can be investigated with compatible representations.

\subsection{From tabular features to typed questions}
The online implementation treats Jev answers as acquired observations with measured costs. Unlike fixed tabular values, these observations can vary across calls. Training obtains answers in batches, whereas sequential testing requests one question at a time; this change in question context may also change answers. Because each policy makes independent live calls, holding the downstream predictor fixed does not completely separate acquisition-policy effects from Jev variability. Repeated requests under matched question contexts would be needed to quantify that component.

The eight questions per task are hand-specified, and binary thresholding discards information in the returned probabilities. Poor performance can therefore arise from question coverage, answer errors, limited training examples, the masked predictor, or the acquisition policy. The direct-choice reference changes both representation and prediction rule. It should not be interpreted as isolating any one of these mechanisms. Adaptive stopping, calibrated uncertainty, question-set learning, and larger independently sampled evaluations remain open directions.

\section{Conclusions}
JevForest turns conditional tree paths into proposals about which feature to observe next, including semantic questions answered by real Jev calls. Global information-gain weighting combines these proposals, and a shared masked predictor maps acquired answers to a label. Offline MiniBooNE results show that weighting effects depend on budget. The online pilot shows task-dependent acquisition performance and a practical limitation: four sequential forest queries are less accurate, more costly, and slower than an eight-question batch on both sampled tasks. Direct Jev classification matches batch accuracy at lower cost. The separate boosting study remains preliminary and does not evaluate a combined system. Path voting is therefore a testable acquisition mechanism, but the present evidence does not establish a deployment advantage; larger studies must evaluate information selection together with answer quality, prediction, and actual query costs.

\section*{Acknowledgments}
OpenAI Codex, a generative AI tool, assisted with literature review, implementation, experimental execution and analysis, and manuscript drafting and revision. It also supported figure preparation and LaTeX typesetting.

\end{document}